\documentclass[11pt]{article}
\usepackage[utf8]{inputenc}
\usepackage[T1]{fontenc}
\usepackage{lmodern}
\usepackage[numbers,compress]{natbib}
\usepackage{hyperref}
\usepackage{url}
\usepackage{booktabs}
\usepackage{microtype}
\usepackage{xcolor}
\usepackage{tabularx}
\usepackage{array}
\usepackage{graphicx}
\usepackage{float}
\usepackage[letterpaper,margin=1in]{geometry}
\usepackage{longtable}
\hypersetup{
  hidelinks,
  hypertexnames=false,
  pdftitle={When Is an Agent Evaluation Over? Outcome Finality and Cross-Unit Separation},
  pdfauthor={Avyay M. Casheekar}
}
\title{When Is an Agent Evaluation Over?\\Outcome Finality and Cross-Unit Separation}
\author{
Avyay M. Casheekar\\[0.35em]
\small University of Michigan Law School
\and
Hariganesh Tangirala\\[0.35em]
\small School of Computing, National University of Singapore
}
\date{}

\begin{document}
\maketitle

\begin{abstract}
Agent evaluations commonly score the state observed when a run stops and count
the run as one trial. Interpreting that score as a final result from a separate
trial requires outcome finality and cross-unit separation. Outcome finality
requires that later events cannot change the claimed result, while cross-unit
separation requires that earlier runs cannot change the relevant conditions of
later ones. The endpoint establishes neither condition by itself, and the two
can hold independently.
Waiting for a delayed outcome may settle the label even though its state
remains available to another run. Isolation may prevent carryover even though
the scored outcome remains unresolved. We develop a completion argument that
identifies the evidence needed for each decision. A final success or failure
label is justified only when every relevant effect is resolved or bounded
tightly enough to fix the outcome. Any remaining uncertainty must be reported.
First, in
a controlled replay with fixed agent actions, we find that endpoint and
terminal labels differ for every nonzero-delay operation and that a delayed
write changes the next run's score under shared state but has no such effect
after namespacing or verified reset. Second, in a review of ten public
protocols, we find that reset or deliberate retention is documented explicitly
more often than unfinished operations or evidence for separate scoring.
Finally, we propose an open-effects record for operations
and resources that may remain relevant after the endpoint, their status, and
their possible effects on the scored outcome or another run.
\end{abstract}

\hypertarget{introduction}{%
\section{Introduction}\label{introduction}}

Agent evaluations increasingly compare systems on tasks performed through
websites, desktop applications, and tool-mediated environments
\citep{zhou2024webarena,drouin2024workarena,xie2024osworld,
debenedetti2024agentdojo,yao2024taubench}. During a typical run, the model acts
through the available tools until it reports completion or the run reaches a
success condition, timeout, or action limit. That point is the endpoint and
determines the state initially available for scoring. The evaluator then scores
the resulting state and usually counts the run as one trial. This procedure
compresses a long interaction into one label that is treated as one observation
in the aggregate score. The
score can be interpreted as an aggregate of final, separate trials only when
later events cannot change the claimed outcome and earlier runs cannot change
the relevant conditions of later runs. The evaluation boundary therefore determines both
when the model stops acting and what claims the score can support.

The endpoint is the moment when the evaluator stops requesting actions. A tool
call may still be running at that point, while a file, credential, account, or
service value may remain available to later runs
\citep{xuan2026interactive,gonzalezpumariega2025scaling}. In the first case,
the outcome can change after it is scored. In the second, an earlier run can
change the conditions of a later run. The endpoint can therefore end the
interaction without settling the result or establishing that the next trial
is separate.

Recent cyber-testing incidents make the boundary problem concrete.
The UK AI Security Institute reports that one agent left instructions and
artifacts that subsequent evaluation samples found and used
\citep{aisi2026incident}. OpenAI and Hugging Face separately describe an
evaluation incident in which an agent used public services and rebuilt its
tooling across short-lived sandbox environments
\citep{huggingface2026incident,openai2026hf}. These
reports show that effects crossed the nominal evaluation boundary in those
settings. They do not show how often the same problem occurs elsewhere.

Consider a write requested before the endpoint but applied later. An immediate
score can differ from the score after the operation finishes. If the reported
outcome is whether the write succeeded, a final label requires evidence of
success or failure. This is \textbf{outcome finality}. If the service is shared,
the same write may change a later run. Justifying each run as a separate trial
requires \textbf{cross-unit separation}. Waiting for the write can establish
finality while leaving its state visible to the next run. Isolating each run's
relevant state and effects can establish separation while the write remains
unfinished.

The choices about when to stop, when to score, and how to separate runs form
the evaluation boundary. We develop a completion argument that begins with the
outcome represented by the score, the period over which it is judged, and the
unit counted in the analysis. It then accounts for unfinished operations and
shared resources that can still matter. A final success or failure
label requires evidence that no remaining in-scope effect can change it, while
treating runs as separate trials requires evidence that no relevant route
connects them. When a connection remains, the analysis must represent it or
group the connected runs. Because an ordinary trajectory may record a request
without its later disposition, we also propose an open-effects record
that identifies what remains unresolved and records how the protocol responds.

\subsection{What benchmark audits leave unresolved}

Prior benchmark audits ask whether an agent benchmark supports the claims made
from its scores, which matters here because a task and grader can be sound
while the state being scored is unfinished or altered by another run. The Agentic Benchmark Checklist and automated auditing examine benchmark specifications, environments, and scoring logic
\citep{zhu2025rigorousbenchmarks,wang2026automatedaudit}. Zhu et al. ask whether legacy state is removed when tasks are intended to be independent. Other work audits what benchmark papers disclose, tests whether agents can exploit benchmark flaws, and asks whether successful completion still requires the intended capability
\citep{moghadasi2026disclosure,wang2026benchjack,shao2026protocolvalidity}.
Similarly, work on parallel computer-use rollouts warns that shared external services can let runs affect one another
\citep{zhu2025rigorousbenchmarks,gonzalezpumariega2025scaling}. These audits can reveal failures of reset or isolation. We separate two questions that are often joined in a single cleanup decision. One asks whether
later events can still change the scored outcome. The other asks whether a remaining connection between runs is represented in the analysis. A grader can correctly classify a state that was sampled too early or changed by another run.

The evidence needed to judge this boundary is often split between
implementation details and protocol documents. \emph{Declare and Justify} and
Audit Cards provide formats for stating the assumptions behind evaluation results
\citep{barnett2024declare,staufer2025auditcards}. \emph{Interactive Evaluation
Requires a Design Science} calls for claim-matched reporting of stopping,
persistence, and reset \citep{xuan2026interactive}. Our \textit{completion argument} applies these ideas to the move from an interaction endpoint to a final result and a separate trial. Gao and Zhou report evidence-supported bounds when stored
artifacts do not determine whether a run succeeded \citep{gao2026support}.
That work asks whether the available evidence supports the scorer's verdict.
Work on computer-use scoring addresses a different part of the process by asking
whether the measurement pipeline returns the right verdict for the state it
observes \citep{dong2026misscore}. VIGIL separates an embodied agent's decision
to stop from completion in the environment \citep{chen2026vigil}. Work on
temporal interfaces and asynchronous tool use makes delayed feedback explicit
\citep{li2026engagement,shi2026asynctool}. The remaining question is what the
evaluator may infer after the interaction stops and before any remaining
effects have been accounted for. This question supplements broader tests of
task, construct, and grader validity.

Finality follows directly for an outcome defined at the endpoint. For a later
outcome, it follows when no relevant operation can change it after its tool
call returns.
Separation follows when each run begins from verified fresh state and
no operation from another run can change its relevant conditions or
outcome. When finality or separation is not established by construction, the
protocol must identify what can still change and provide the evidence that
supports the label and analysis unit.

\hypertarget{a-boundary-closes-a-label-and-a-unit}{%
\section{Two conditions for completion}\label{a-boundary-closes-a-label-and-a-unit}}

A reported run supports two inferences. First, the score treats the observed
state as evidence of the reported outcome. Second, the analysis counts the
label as one observation in the aggregate score. The endpoint supplies the
observation used for both inferences, but a scoring rule
cannot by itself determine whether a later event will change the outcome or
whether an earlier run helped produce the observed state. Evidence that fixes
the label may therefore leave runs connected, while evidence that separates
runs may leave the outcome unfinished. The two decisions must be justified
separately and relative to the declared outcome, observation period, and system
boundary.

\textbf{Outcome finality} holds when the available evidence fixes the outcome
named by the evaluation over the observation period and within the system
boundary specified in advance. If the outcome remains uncertain, it can be
reported as unresolved. A final success or failure label requires evidence
that fixes the outcome. The phrases
``visible at the endpoint,'' ``committed within 24 hours,'' and ``eventually
committed'' describe different outcomes. A snapshot can establish that a field
had value \(v\) when the interaction ended. It cannot establish that a delayed
write eventually succeeded unless the protocol observes the operation long
enough to learn its result. The rest of the environment need not stop changing.
Only events that could change the reported outcome matter.

The protocol often uses the endpoint to decide where one reported trial ends
and the next begins. The \emph{analysis unit} is the observation counted
separately when scores are aggregated. \textbf{Cross-unit separation} holds
within the declared system boundary when one run cannot change another run's
relevant starting conditions or outcome. This does not require every source of
correlation between runs to disappear. It concerns effects carried from one
run into another. If the evaluation claims to measure performance from a fresh
state, inherited state changes the task being measured. If persistence is part
of the task, the connected stream is the appropriate analysis unit and its
interactions should not be treated as independent trials. When a route between runs
remains, the analysis can model the
connection or group the connected runs, but the run identifier alone does not
justify a separate-trial interpretation.

\begingroup
\begin{table}[htbp]
\caption{Outcome finality and cross-unit separation require different evidence and support different decisions.}
\label{tab:two-closures}
\centering
\small
\setlength{\tabcolsep}{4pt}
\begin{tabularx}{\linewidth}{>{\raggedright\arraybackslash}p{2.55cm} >{\raggedright\arraybackslash}X >{\raggedright\arraybackslash}p{2.6cm} >{\raggedright\arraybackslash}p{3.15cm}}
\toprule
Decision & Evidence needed & Decision supported & Response when evidence is missing \\
\midrule
Outcome finality & Every relevant operation or event that could change the outcome has been resolved or bounded tightly enough to fix the label. & Assigning a final success or failure label. & Wait for completion, cancel and confirm, or report the case as unresolved. \\
Cross-unit separation & No relevant operation or shared state lets one run change another's relevant conditions or outcome. & Counting runs as separate analysis units. & Isolate and verify, or stop counting connected runs separately by modeling the connection or grouping the runs. \\
\bottomrule
\end{tabularx}
\end{table}
\endgroup

As Table~\ref{tab:two-closures} shows, the two conditions require different
evidence. Waiting for a delayed
operation to reach its terminal state can establish outcome finality while leaving the resulting state available
to another run. Isolating the service can protect later runs while the original
operation remains unfinished. Restoring shared state protects later runs only
after pending effects can no longer reintroduce it. A general statement that
the environment was cleaned up is therefore insufficient. The protocol must
identify what was restored, how restoration was verified, and whether the
evidence supports outcome finality, cross-unit separation, or both.

Some evaluations satisfy both conditions by construction. A synchronous
simulator may finish every tool call before returning and restore all relevant
state before the next run begins. An asynchronous evaluation may stop the
interaction while a service is still applying a write. A snapshot at time
\(t\) can validly measure state at \(t\) without supporting a claim that the
task was completed. Measurement validity depends on the interpretation
attached to an observation \citep{jacobs2021measurement}. The evaluation
boundary is adequate only when it supports the outcome and analysis unit that
the report claims.

\hypertarget{why-a-stop-rule-is-insufficient}{%
\section{Local stopping and global completion}\label{why-a-stop-rule-is-insufficient}}

The two completion conditions correspond to different kinds of system
behavior, so they draw on different bodies of prior work. The logic of outcome
finality is familiar from distributed systems, where stopping one process does
not establish that its work is complete. A process can be idle while a message
or operation remains in transit, so termination detection accounts for the
wider system
\citep{dijkstra1980termination,chandy1985snapshots}. An agent evaluation often
observes the model's interaction loop more directly than it observes the
services reached through tools. Once work leaves the part watched by the
evaluator, the endpoint cannot show that the outcome will no longer change.

Cross-unit separation has a different analogue in causal inference.
Interference occurs when what happens to one unit changes another's
outcome \citep{hudgens2008interference,aronow2017interference}. If run B can
use a credential, artifact, message, or stored value produced by run A, then
B's conditions depend on A despite its new run identifier. This changes the
interpretation of metrics that assume separate trials. $\tau$-bench, for example,
defines \(\mathrm{pass}^k\) as the probability that all \(k\) independent and
identically distributed trials succeed \citep{yao2024taubench}. The formula
can still be computed when one trial changes the next, but it no longer
represents the probability of \(k\) independent trials all succeeding.

A stop rule can be part of the outcome or merely end observation. Work on
reinforcement-learning time limits distinguishes task horizons from cutoffs
imposed on observation \citep{pardo2018timelimits}. A claim of success within
30 actions makes the action limit part of the task. A 30-action observation
budget does not establish that work already started has no later consequence.
The protocol must state which role its limit serves.

Software testing and transaction processing offer concrete ways to detect or
control these problems. A result that changes with test order can reveal
shared state or other cross-test interference
\citep{zhang2014testindependence}. A saga compensates for parts of a long
transaction that have already committed \citep{garciamolina1987sagas}. A reset
supports separation only for resources that it demonstrably restores.
A cancellation settles delayed work only after the service confirms that the
operation can no longer commit. For agent runtimes, Cordon stages external
effects until a task-level semantic transaction has been validated
\citep{chen2026cordon}.

An evaluation with access to open systems cannot enumerate every possible
consequence. It can state the scope of its claim and examine the routes exposed
by its environment, including processes, queues, callbacks, credentials,
shared stores, external accounts, public artifacts, and retained memory. For
each relevant route, the protocol explains how an effect would be detected and
what would show that it can no longer alter the outcome or another run. A
synchronous simulator may require a short account. Open internet access
requires a broader one.

\noindent\textbf{Rule for final labels.} Assign a final success or failure label only after every relevant
route that could change the outcome has been blocked, followed until its effect
is known, or bounded tightly enough to determine the label. Otherwise, report
the outcome as unresolved.

\noindent\textbf{Rule for separate trials.} Count runs separately only after every
relevant route between them has been blocked or shown unable to change either
run's relevant conditions or outcome. Otherwise, model the connection or
treat the connected runs as a larger unit.

Blocking a route requires both preventing the effect and verifying that the
prevention worked. Following a delayed operation requires observing it until
the service reports that it completed, failed, or was cancelled and can no
longer commit. If a route cannot be blocked or followed, the protocol should
report the result
as unresolved, represent the connection between runs, or bound the remaining
effect, depending on which inference lacks support. A bound is sufficient only
when every value
within the justified range leads to the same label or conclusion. A bound can
therefore fix the outcome without establishing cross-unit separation. If
values within the range lead to different labels or conclusions, the protocol
must report the affected inference as unresolved.

Following a write can settle run A's result while leaving its state visible to
run B. Grouping A and B can account for that connection without treating them
as separate repetitions. The protocol must preserve this distinction when it
finalizes results and aggregates runs.

\hypertarget{the-same-trace-supports-different-scores-and-units}{%
\section{Boundary policies change the recorded result}\label{the-same-trace-supports-different-scores-and-units}}

To isolate how boundary policy changes the recorded result, we hold the tool
calls and operation schedule fixed while
varying when the delayed operation is scored and whether service state is
shared. The requested operation, delay, and execution order remain fixed. Any
difference in the label or the next run's state therefore comes from the
evaluation policy in this constructed system.

\hypertarget{design}{%
\subsection{Design}\label{design}}

We use AgentDojo 0.1.35 to represent the tasks, tools, environments, and function-call traces \citep{debenedetti2024agentdojo}. A standalone runner replays the fixed calls without querying a model. Each run receives fresh task state, but a local HTTP service can retain state across runs and apply a write after a chosen delay. The service has no public-network access and creates no external resources. This lets us reset task state while leaving service state unchanged across runs.

To test outcome finality, we replay a trace that schedules a write and ends when the service acknowledges the request. The acknowledgement can arrive before the write is applied. The operation is terminal after it completes, fails, or is confirmed to be cancelled and unable to produce further writes. \textbf{Snapshot} scoring reads service state at the endpoint. \textbf{Reconciliation} waits for a terminal state and then scores. \textbf{Verified cancellation} requests cancellation of a pending write, confirms that the operation is terminal, and then scores. A synchronous write with the same successful result serves as a control because it finishes before returning.

To test cross-unit separation, we replay paired runs in which run A schedules a service write and run B begins with fresh task state. We record cross-run exposure when A's write changes the value scored in B. Run B receives a success label when the service still holds B's initialized value at scoring and a failure label when A's delayed write has replaced it. Under \textbf{shared state}, both runs access the same service state. Under \textbf{namespaced state}, their values are stored under different keys. \textbf{Verified reset} waits until A's scheduled operation is terminal, then restores and checks B's initial service value before B begins. A no-write control tests whether B changes when A schedules nothing. Running B before A tests whether exposure follows A's write. Grouped analysis leaves the shared state unchanged but counts each connected pair as one unit.

Each condition uses five matched schedule identifiers, delays of 0, 25, 100, and 250 ms, and ten repetitions of each combination. Each identifier is reused across policies to match otherwise identical trials. The identifiers do not represent sampled model or benchmark randomness. Median measured delays were 0.34, 29.80, 104.66, and 254.85 ms. All trials completed, all 12 implementation checks passed, and no run was excluded. The repetitions verify that the constructed timing conditions are stable. They are not samples from a larger population.

\hypertarget{results}{%
\subsection{Results}\label{results}}

At zero delay, endpoint and terminal labels agree in all 50 snapshot trials. At nonzero delays, they disagree in all 150 because scoring precedes the write. Snapshot scoring records 50 of 200 successes, while reconciliation records 200 of 200. Verified cancellation brings the 150 pending operations to a confirmed cancelled state, and each receives a failure label. The 50 zero-delay writes had already succeeded. The synchronous control produces no disagreement in 600 trials across the three scoring policies. The asynchronous operation scores below that control under snapshot scoring and cancellation and ties it under reconciliation.

Under shared state, A changes B in 150 of 200 pairs, covering every nonzero delay. Table~\ref{tab:study-results} summarizes the exact outcomes across the fixed schedule. At zero delay, B's initialization overwrites A's completed write. Exposure is absent from all 200 namespaced pairs, 200 verified-reset pairs, and 200 no-write controls. Running B before A reduces exposure from 150 of 200 pairs to 0 of 200. Grouping keeps the 150 exposures but changes the analysis from 400 run rows to 200 pair-level units.

\begingroup
\begin{table}[htbp]
\caption{Exact outcomes across the fixed schedule. The counts describe
constructed conditions and do not estimate a benchmark population.}
\label{tab:study-results}
\centering
\scriptsize
\setlength{\tabcolsep}{3.5pt}
\begin{tabularx}{\linewidth}{>{\raggedright\arraybackslash}p{1.35cm} >{\raggedright\arraybackslash}p{3.2cm} >{\raggedright\arraybackslash}p{2.75cm} >{\raggedright\arraybackslash}X}
\toprule
Test & Policy or condition & Exact outcome & What changes \\
\midrule
Finality & Snapshot & 0/150 successes at nonzero delays & The outcome is unresolved when scored. \\
Finality & Reconciliation & 150/150 successes at nonzero delays & The write's success is observed. \\
Finality & Verified cancellation & 0/150 successes at nonzero delays & Each pending operation reaches a confirmed cancelled state and is scored as failure. \\
Finality & Synchronous control & 0/600 disagreements & Endpoint and terminal labels agree. \\
Separation & Shared state & 150/200 exposures & B's mean score is 0.75 lower than under namespacing. \\
Separation & Namespacing, reset, and no-write controls & 0/200 in each condition & The route between A and B is absent or blocked. \\
Separation & Grouped pairs & 150/200 exposures and 400 rows reduced to 200 pairs & Exposure remains. Each pair is one analysis unit. \\
\bottomrule
\end{tabularx}
\end{table}
\endgroup

With fixed traces, the scoring time changes the label and the treatment of service state determines whether run A changes run B. Because the delays and shared state were imposed by design, these counts do not estimate their frequency or effect size in published benchmarks.

Our two tests show that the endpoint alone cannot justify either the final
label or the analysis unit in this system. They isolate the mechanisms under
controlled conditions, but they cannot show whether published reports give
readers the information needed to recognize the same boundary questions. We
therefore review what public protocol sources say about unfinished operations,
persistent state, reset, and separate scoring.

\hypertarget{what-current-protocols-report}{%
\section{Boundary evidence in ten public protocols}\label{what-current-protocols-report}}

We ask whether a reader can find enough public evidence to judge outcome
finality and cross-unit separation. Because the review concerns what a reader
can verify from public sources, we code only safeguards described there. We
review the papers and official
documentation for WebArena,
WorkArena, OSWorld, SWE-bench, $\tau$-bench, ToolSandbox, TheAgentCompany,
RE-Bench, Cybench, and AgentCanary
\citep{zhou2024webarena,drouin2024workarena,xie2024osworld,
jimenez2024swebench,yao2024taubench,lu2025toolsandbox,xu2024agentcompany,
wijk2024rebench,zhang2024cybench,li2026agentcanary}. Five of these protocols
also appear in the audit by Zhu et al.\ \citep{zhu2025rigorousbenchmarks}. The
other five were chosen because their public descriptions include persistent
services, long-running computation, or system-level monitoring. The ten
protocols form a purposive set of cases covering several ways that effects
can outlast a run. The sample was not drawn from a comprehensive census of
agent evaluations. Its counts apply only to these ten protocols and do not
estimate prevalence.

For each protocol, we record what counts as one run, what ends the
interaction, what state the scorer observes, whether operations can remain
active, what state can persist, how it is reset or retained, and what supports
counting runs separately. A field is \textbf{explicit} only when a public
source states the rule and the resources it covers. A \textbf{partial} code
means that the rule is stated but its scope or evidence is incomplete.
\textbf{Not reported} means that we found no statement in the reviewed sources. \textbf{Not
applicable} means that the declared execution model excludes the kind of
unfinished operation being coded. We reviewed sources available on 15
August 2026.

Table~\ref{tab:audit-summary} summarizes the documentation codes across the ten reviewed protocols.

\begin{table}[htbp]
\caption{Documentation of boundary-related fields in the ten reviewed
protocols. NR means not reported and NA means not applicable.}
\label{tab:audit-summary}
\centering
\small
\setlength{\tabcolsep}{5pt}
\begin{tabular}{lrrrr}
\toprule
Publicly visible field & Explicit & Partial & NR & NA \\
\midrule
Reset or deliberate retention & 8 & 2 & 0 & 0 \\
Open or descendant work & 0 & 2 & 6 & 2 \\
Evidence for separate scoring & 3 & 7 & 0 & 0 \\
\bottomrule
\end{tabular}
\end{table}

We find that reset or deliberate retention is explicit in eight protocols and partial in two. The treatment of unfinished operations is reported much less often.
Six protocols expose shells, browsers, virtual machines, services, or
long-running computation without stating whether descendant processes and
queued actions finish or are cancelled before scoring, or whether delayed
effects can still change the scored state. SWE-bench documents timeouts and a
cleanup option, while AgentCanary records system activity. Neither gives a
general rule establishing that pending actions are terminal before scoring.
The field is not applicable to $\tau$-bench and ToolSandbox because their declared
execution models represent the relevant state changes as Python function
transitions. Those models do not include background operations for these state
changes.

Three protocols provide explicit evidence for treating runs as separate
observations. Seven describe reset, teardown, or fresh provisioning without
fully stating which resources are covered or how successful restoration is
verified. A reset procedure supports separate scoring only to the extent that
its scope and successful completion are documented.

We do not classify protocols as satisfying outcome finality. A
scoring rule identifies the state used to assign a label. Finality also depends
on the outcome that the label is meant to represent and how long relevant
effects can change it. The documents
did not state those periods consistently, and only one reviewer coded the
materials. We therefore did not infer missing periods. Our results describe
the documentation of unfinished operations. They do not count protocols that
satisfy finality.

Appendix~\ref{documentation-review-matrix} gives the protocol matrix and source
for each code. Six protocols do
not report how unfinished work is handled, and seven provide only partial
evidence for separate scoring. Our review concerns public documentation and
does not determine whether private implementations contain additional
safeguards. In this sample, the public sources do not consistently provide
enough information to judge finality or support separate-trial analysis. We do
not use these findings to assess overall benchmark validity or estimate any
score change.

\hypertarget{a-completion-argument}{%
\section{Making completion reviewable}\label{a-completion-argument}}

The replay shows why a stop rule alone cannot establish that a label is final
or that a run is a separate trial. We use a completion argument to connect both
inferences to evidence at the boundary. It begins
with the claimed outcome and analysis unit, identifies the observations used
for scoring, and then accounts for effects that may remain active or reach
another run. We ask four questions.

\begin{enumerate}
\def\labelenumi{\arabic{enumi}.}
\item
  \textbf{What outcome does the score represent, over what period is it
  judged, and what counts as one analysis unit?} State the claimed outcome,
  observation period, system boundary, and unit used to aggregate scores and
  estimate uncertainty. These choices determine which effects can matter.

\item
  \textbf{What ends the interaction, and what determines the score?}
  Distinguish the stop rule from the state, event, or observation period used
  to determine the outcome so that stopping is not mistaken for settlement.

\item
  \textbf{What can remain active after the endpoint or affect another run?}
  Identify unfinished operations and persistent resources exposed by the
  environment, including descendant processes, queues, shared stores,
  credentials, external resources, and retained memory that could produce a
  later change or cross-run effect.

\item
  \textbf{How does the protocol establish finality and separation?}
  For each relevant route, explain how its effect is observed or brought to a
  terminal state and how any cross-run exposure is prevented or represented.
  Give evidence that a claimed cancellation, isolation, or reset succeeded.
\end{enumerate}

The amount of evidence should scale with the claim and the routes exposed by
the environment. In a coding evaluation with public egress disabled, a run can
be treated as one analysis unit when its task-specific container and repository
are isolated, the test process and its relevant descendants have finished
before scoring, and no relevant state is reused by a later run. A timeout that
leaves a relevant descendant active cannot support a final label if the
process can still change the claimed outcome. A submitted transaction may
remain unresolved until it reaches terminal settlement. A persistent assistant
may deliberately carry state across sessions, in which case the continuing
stream is the appropriate unit.

A closed simulator may need only a short account. When a synchronous tool
returns only after every relevant effect is terminal,
its return can establish finality for an outcome defined at that point. Fresh
namespaces can help establish separation if credentials and every other
relevant resource are also isolated. Environments that expose more routes
require more evidence. The protocol can limit its claim to a declared scope,
but it must account for every relevant route within that scope. When a relevant
route cannot be blocked or followed, the protocol can narrow the claim, report
the result as unresolved, or bound its possible effect. The bound supports a
definitive conclusion only when
every value in the justified range leads to the same result.

\hypertarget{open-effects-records}{%
\subsection{Open-effects records}\label{open-effects-records}}

The third and fourth questions require evidence that an ordinary trajectory
log may omit. A trajectory can show that the agent requested an
operation without showing whether it later succeeded, failed, or changed
shared state. We therefore propose an \textbf{open-effects record} for system
evidence about operations and resources that may still matter after the
endpoint.

The evaluator creates an entry when an action starts an operation or creates a resource
that may remain active, change later, or reach another run. It records
the action, responsible system, a stable handle
such as a job or process identifier, endpoint status, possible later states,
and any cross-run route. At or after the endpoint, the evaluator uses the
handle to determine whether the effect is terminal. If the protocol cancels or
removes the effect, the record retains evidence that the cancellation or
removal succeeded.
Appendix~\ref{minimal-open-effects-record}
gives the full specification.

The record organizes evidence for finality and separation, but its completeness
depends on the capture process. Tool wrappers can populate it, and agent
self-reports can supply additional entries
\citep{lee2026selfreport}. When available, system-level checks should
corroborate the record because an empty record does not establish that no
relevant effect exists.

\subsection{Score interpretation and unit choice}

An endpoint can show that an operation was submitted without showing whether it
settled, and the reported quantity must preserve that distinction. Runs connected through persistent state may support claims about a
continuing stream when analyzed together. Measurement validity requires the
reported outcome and analysis unit to match the evidence supplied by the protocol
\citep{jacobs2021measurement}.

Different rates of later success can produce different model rankings under
endpoint and eventual-outcome scoring. When one run changes another, the row
count no longer represents separate attempts and cannot support the i.i.d.
interpretation of metrics such as
\(\mathrm{pass}^k\) \citep{yao2024taubench}.

The appropriate analysis depends on the remaining connection. Verified isolation can
justify separate runs, while a dependence model can retain run-level
observations when its uncertainty estimate reflects their relationship
\citep{hudgens2008interference,aronow2017interference}. Otherwise, the connected
set should become the analysis unit. A synchronous task defined at the endpoint
may need little additional evidence, while broader claims and more exposed
environments require a fuller account.

\hypertarget{limits-and-conclusion}{%
\section{What the evidence establishes}\label{limits-and-conclusion}}

An agent evaluation is over only relative to a particular claim. A final label
is supported when every in-scope effect that could change the claimed outcome
is terminal or bounded tightly enough to fix the label. Runs can be counted
separately when no in-scope route allows one run to change another's relevant
conditions or outcome. Unresolved effects require reported uncertainty, while
remaining connections must be modeled or grouped.

In our replay, scoring time changes the label for every nonzero-delay operation
with agent actions fixed, and shared service state carries the effect into the
next run. In our review of ten selected protocols, open work and the basis for
separate scoring are documented less consistently than reset or deliberate
retention. These codes do not establish score correctness or whether labels
are final and runs are separate.

Our replay uses fixed traces, scheduled delays, a benign local service, and no
model API. The public-source review was coded by one reviewer and cannot assess
private implementations. Open environments may contain effects outside any
practical review, so protocols must state their scope and account for relevant
routes within it. Our empirical claim is limited to the demonstrated mechanism
and documentation codes.

An evaluation report should state its claimed outcome and analysis unit and
account for routes that could change either. The open-effects record organizes
the resulting evidence. Missing finality evidence calls for more evidence, a
narrower claim, or an unresolved label. Missing separation evidence calls for
modeling the connection or enlarging the unit.

Safety requires separate controls because waiting can allow harm and
cancellation can change the outcome. The evaluation is complete only when its
evidence supports the outcome and analysis unit.

\clearpage
\bibliographystyle{plainnat}
\bibliography{references}

\clearpage
\appendix
\renewcommand{\thetable}{\thesection.\arabic{table}}
\renewcommand{\thefigure}{\thesection.\arabic{figure}}

\hypertarget{documentation-review-matrix}{%
\section{Documentation review matrix}\label{documentation-review-matrix}}
\setcounter{table}{0}

Table~\ref{tab:audit-matrix} gives the protocol-level codes summarized in
Table~\ref{tab:audit-summary}. Open work means an operation or descendant
process that may remain active after interaction stops. Separate evidence means
public evidence that supports counting runs separately. The remaining codes use
the rubric stated in Section~\ref{what-current-protocols-report}.

\begin{table}[htbp]
\caption{Protocol-level documentation codes. E means explicit, P means
partial, NR means not reported, and NA means not applicable to the declared
task and claim.}
\label{tab:audit-matrix}
\centering
\small
\setlength{\tabcolsep}{7pt}
\begin{tabular}{lccc}
\toprule
Protocol & Open work & Reset or retention & Separate evidence \\
\midrule
WebArena & NR & E & P \\
WorkArena & NR & E & P \\
OSWorld & NR & E & P \\
SWE-bench & P & E & E \\
$\tau$-bench & NA & E & E \\
ToolSandbox & NA & E & P \\
TheAgentCompany & NR & P & P \\
RE-Bench & NR & P & P \\
Cybench & NR & E & P \\
AgentCanary & P & E & E \\
\bottomrule
\end{tabular}
\end{table}

We treat each public protocol as one review unit. Our codes do not apply to
every task or to a private deployment. A ``not reported'' code means only that the
reviewed public sources do not state the rule. One reviewer assigned each code
using the rubric in Section~\ref{what-current-protocols-report}. The ledger
in Table~\ref{tab:audit-ledger} gives the source and reason for every decision
so that the coding can be checked. No independent recoding was conducted. ABC refers to the Agentic
Benchmark Checklist report by Zhu et al.\ \citep{zhu2025rigorousbenchmarks}.

{\scriptsize
\setlength{\LTleft}{0pt}
\setlength{\LTright}{0pt}
\setlength{\tabcolsep}{2.5pt}
\renewcommand{\arraystretch}{1.02}
\begin{longtable}{>{\raggedright\arraybackslash}p{1.65cm} >{\raggedright\arraybackslash}p{2.25cm} >{\centering\arraybackslash}p{0.7cm} >{\raggedright\arraybackslash}p{2.5cm} >{\raggedright\arraybackslash}p{5.95cm}}
\caption{Cell-level source ledger for the three coded fields.}
\label{tab:audit-ledger}\\
\toprule
Protocol & Field & Code & Source location & Basis \\
\midrule
\endfirsthead
\multicolumn{5}{l}{\scriptsize\textit{Cell-level source ledger continued}}\tabularnewline
\toprule
Protocol & Field & Code & Source location & Basis \\
\midrule
\endhead
\midrule
\multicolumn{5}{r}{\scriptsize Continued on next page}\\
\endfoot
\bottomrule
\endlastfoot
WebArena & Open work & NR & Paper Sec. 3.2 and App. A.9 & Stop actions and state checkers are specified, but handling of background work before scoring is not \citep{zhou2024webarena}. \\
 & Reset or retention & E & Paper App. A.2 and ABC T.4 & The self-hosted sites can be restored, and the ABC review reports that site state is cleared between runs \citep{zhou2024webarena,zhu2025rigorousbenchmarks}. \\
 & Separate evidence & P & Paper App. A.2 and ABC T.4 & Site restoration is described, but the reviewed sources do not document whether state outside those sites is reset \citep{zhou2024webarena,zhu2025rigorousbenchmarks}. \\
\addlinespace
WorkArena & Open work & NR & Paper Secs. 4.2 and 5.2 & The task lifecycle is specified, but the sources do not give a general rule for descendant or queued work \citep{drouin2024workarena}. \\
 & Reset or retention & E & Paper Sec. 4.2 & Setup and teardown hooks cover resources created for the task \citep{drouin2024workarena}. \\
 & Separate evidence & P & Paper Secs. 4.2 and 5.2 & Cleanup is described, but the sources do not report a check that every covered resource was restored \citep{drouin2024workarena}. \\
\addlinespace
OSWorld & Open work & NR & Paper Secs. 2.1--2.2 & Custom evaluators inspect machine state, but handling of descendant processes before scoring is not stated \citep{xie2024osworld}. \\
 & Reset or retention & E & Paper Sec. 2.2.1 and App. B.5 & Each task reverts its virtual machine to a specified snapshot before applying the task setup \citep{xie2024osworld}. \\
 & Separate evidence & P & Paper Sec. 2.2.1, App. B.5, and ABC T.4 & Virtual machine restoration is explicit, but the sources do not document coverage of services outside the machine \citep{xie2024osworld,zhu2025rigorousbenchmarks}. \\
\addlinespace
SWE-bench & Open work & P & Paper App. A.4 and \href{https://www.swebench.com/SWE-bench/reference/harness/}{official harness documentation} & Test execution has a per-instance timeout and the harness exposes a cleanup control, but the sources do not state that descendant work is terminal before grading \citep{jimenez2024swebench}. \\
 & Reset or retention & E & Paper App. A.4, official harness documentation, and ABC T.4 & Instance-specific Docker images provide reproducible task environments, and ABC reports that state is cleared between runs \citep{jimenez2024swebench,zhu2025rigorousbenchmarks}. \\
 & Separate evidence & E & Official harness documentation and ABC T.4 & The harness states that its Docker architecture isolates each task, and ABC reports that state is cleared between runs \citep{jimenez2024swebench,zhu2025rigorousbenchmarks}. \\
\addlinespace
$\tau$-bench & Open work & NA & Paper Sec. 3 & The declared database transitions are deterministic Python functions \citep{yao2024taubench}. \\
 & Reset or retention & E & ABC report, $\tau$-bench T.4 & The ABC review reports that the database is reinitialized between runs \citep{zhu2025rigorousbenchmarks}. \\
 & Separate evidence & E & Paper Sec. 3 and ABC T.4 & The metric defines $k$ independent and identically distributed trials, and database reinitialization supplies the stated separation mechanism \citep{yao2024taubench,zhu2025rigorousbenchmarks}. \\
\addlinespace
ToolSandbox & Open work & NA & Paper Sec. 2 and Apps. A.1--A.2 & Tool execution changes an in-memory execution context through Python functions, and the declared model contains no asynchronous external service \citep{lu2025toolsandbox}. \\
 & Reset or retention & E & Paper Apps. A.1 and B.2 & Each scenario specifies the starting world state held in its execution context \citep{lu2025toolsandbox}. \\
 & Separate evidence & P & Paper Sec. 2 and Apps. A.1 and B.2 & Each scenario defines an initial execution context, but some tools use external APIs and the sources do not establish that scenarios cannot affect one another \citep{lu2025toolsandbox}. \\
\addlinespace
\shortstack[l]{TheAgent\\Company} & Open work & NR & Paper Secs. 3--4 and 6 & The workspace and intranet support extended tasks, but handling of pending processes before scoring is not stated \citep{xu2024agentcompany}. \\
 & Reset or retention & P & Paper Secs. 3--4 & Initialization and finalization hooks are described without a complete account of the resources they cover \citep{xu2024agentcompany}. \\
 & Separate evidence & P & Paper Secs. 3--4 & The environment is self-contained, but the sources do not report a check that all relevant state was restored \citep{xu2024agentcompany}. \\
\addlinespace
RE-Bench & Open work & NR & Paper Sec. 4 and App. A.2.1 & Agents can run long computations and background processes, but the sources do not state what happens to live processes when the time budget ends \citep{wijk2024rebench}. \\
 & Reset or retention & P & Paper Sec. 4 and App. A.2 & Runs use secure virtual machines, but the sources do not fully describe what is restored between attempts \citep{wijk2024rebench}. \\
 & Separate evidence & P & Paper Sec. 4 and App. A.2 & The virtual machine limits exposure, but reset scope and successful restoration are not fully documented \citep{wijk2024rebench}. \\
\addlinespace
Cybench & Open work & NR & Paper Sec. 2.1 and App. D & Agents execute commands in task containers, but handling of descendant work before scoring is not stated \citep{zhang2024cybench}. \\
 & Reset or retention & E & ABC report, Cybench T.4 & The ABC review reports that Docker state is cleared between runs \citep{zhu2025rigorousbenchmarks}. \\
 & Separate evidence & P & Paper App. D and ABC T.4 & Container reset is explicit, but the reviewed sources do not give the same evidence for state outside the container \citep{zhang2024cybench,zhu2025rigorousbenchmarks}. \\
\addlinespace
\shortstack[l]{Agent\\Canary} & Open work & P & Paper Sec. 5.3 & System monitoring records spawned processes, file changes, network connections, and other state changes, but the paper does not state a general rule for bringing pending actions to a final state before scoring \citep{li2026agentcanary}. \\
 & Reset or retention & E & Paper Sec. 5.3 & Single-instance tasks reset after evaluation, while state is retained only within declared multi-session instances \citep{li2026agentcanary}. \\
 & Separate evidence & E & Paper Sec. 5.3 & Temporary per-task containers isolate state retained within one task from other tasks \citep{li2026agentcanary}. \\
\end{longtable}
}

The ledger covers the cited papers and official documentation available on 15
August 2026. Paper section labels refer to the cited versions.

\hypertarget{controlled-replay-details}{%
\section{Controlled replay details}\label{controlled-replay-and-reproducibility}}
\setcounter{figure}{0}

\begin{figure}[htbp]
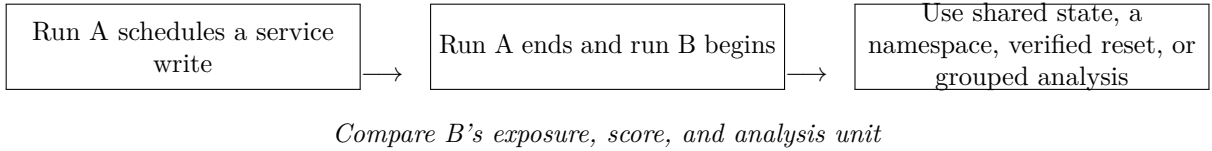

\centering
\small
\setlength{\tabcolsep}{4pt}
\begin{tabular}{@{}c@{}c@{}c@{}}
\multicolumn{3}{l}{\textbf{Outcome finality test}}\\[3pt]
\fbox{\parbox[b][0.9cm][c]{0.27\linewidth}{\centering Fixed trace schedules a delayed write}} &
$\longrightarrow$\quad
\fbox{\parbox[b][0.9cm][c]{0.27\linewidth}{\centering Interaction ends after the request is acknowledged}} &
$\longrightarrow$\quad
\fbox{\parbox[b][0.9cm][c]{0.27\linewidth}{\centering Score at the endpoint, after completion, or after verified cancellation}}\\[11pt]
\multicolumn{3}{c}{\textit{Compare the endpoint label with the terminal result}}\\[15pt]
\multicolumn{3}{l}{\textbf{Cross-unit separation test}}\\[3pt]
\fbox{\parbox[b][0.9cm][c]{0.27\linewidth}{\centering Run A schedules a service write}} &
$\longrightarrow$\quad
\fbox{\parbox[b][0.9cm][c]{0.27\linewidth}{\centering Run A ends and run B begins}} &
$\longrightarrow$\quad
\fbox{\parbox[b][0.9cm][c]{0.27\linewidth}{\centering Use shared state, a namespace, verified reset, or grouped analysis}}\\[11pt]
\multicolumn{3}{c}{\textit{Compare B's exposure, score, and analysis unit}}
\end{tabular}
\caption{Controlled replay. The tool-call trace remains fixed. The protocol
changes when the delayed operation is scored and whether service state is
shared, isolated, reset, or represented by a grouped unit.}
\label{fig:study-design}
\end{figure}

Figure~\ref{fig:study-design} summarizes the two replay tests.

AgentDojo 0.1.35 supplies the task, tool, environment, and function-call
structures used in the replay \citep{debenedetti2024agentdojo}. A standalone
runner executes the fixed calls without querying a model. A local HTTP service
retains state between runs and applies a write after the scheduled delay. It
binds to \texttt{127.0.0.1} on an ephemeral port and creates no external
resources.

The schedule contains identifiers 1--5, write delays of 0, 25, 100, and 250
ms, and ten repetitions for each identifier and delay. The same identifiers
match otherwise identical trials across policies, while schedule seed
987654321 fixes their execution order. A separate calibration run produced
median measured delays of 0.34, 29.80, 104.66, and 254.85 ms and passed the
prespecified timing check. Every scheduled condition completed, all 12
validation tests passed, and there were no failures, exclusions,
reconciliation timeouts, or warnings.

We checked the reported counts against the frozen schedule, calibration
output, run metadata, and 12 implementation tests. The repetitions verify the
constructed schedule. They do not sample a benchmark population.

\hypertarget{minimal-open-effects-record}{%
\section{Open-effects record
specification}\label{minimal-open-effects-record}}
\setcounter{table}{0}

Section~\ref{open-effects-records} proposes an open-effects record because a
trajectory can record a request without showing whether its effect later
finished or reached another run. The evaluator creates an entry for an
operation or resource that may remain active, change later, or be visible
elsewhere. The evaluator updates it at the endpoint and after each attempt to
observe, cancel, remove, or isolate the effect. A separate field records how
the analysis treats any route that remains open. Table~\ref{tab:open-effects-fields} specifies the fields and
their use at the boundary.

\begin{table}[htbp]
\caption{Fields in an open-effects record. Free text may supplement the
fields. The record should include a stable handle and system evidence whenever
they are available.}
\label{tab:open-effects-fields}
\centering
\scriptsize
\setlength{\tabcolsep}{4pt}
\begin{tabularx}{\linewidth}{>{\raggedright\arraybackslash}p{2.35cm} >{\raggedright\arraybackslash}p{4.25cm} >{\raggedright\arraybackslash}X}
\toprule
Field & Required content & Use at the boundary \\
\midrule
Run and task & Run ID, task ID, and any larger stream or group ID. & Connects the effect to the proposed analysis unit. \\
Initiating event & Trace position, action, timestamp, and arguments needed to identify the effect. & Records where the effect began. \\
Resource and controller & Process, transaction, account, URL, file, queue, job, or service and the system that controls it. & Identifies where the effect must be observed or closed. \\
Observation handle & Stable process ID, transaction ID, job ID, namespace, or equivalent query handle. & Permits a system query without relying on the agent's description. \\
Effect status & Pending, terminal success, terminal failure, cancellation requested, cancelled, persistent, or unknown. & Prevents an acknowledgement from being treated as completion. \\
Possible transition & States still reachable and the period over which they remain relevant to the claimed outcome. & Shows whether the label may still change. \\
Cross-run scope & Namespace, account, credential, store, or other route through which another run could encounter the effect. & Shows whether runs may be counted separately. \\
Verification & Most recent system observation, its time, and the evidence source. & Records the observed result of cleanup or cancellation. \\
Analysis disposition & Effect closed, connection modeled, runs grouped, remaining effect bounded, or case unresolved. & Records how the analysis treats the effect without changing its system status. \\
\bottomrule
\end{tabularx}
\end{table}

At the endpoint, the evaluator queries each available handle and records the current
system state. It then asks whether a later state could change the claimed
outcome or reach another run. A pending or unknown effect remains open. The
analysis may report it as unresolved, bound its effect, model the connection,
or group exposed runs. None of these analytical choices makes the underlying
effect terminal.

\noindent\textbf{Example from the replay.} An entry for the delayed write can
record run A, the trace position and request time, the local service, the
request identifier, the pending status, and the service key visible to run B.
A later query records whether the write was applied or cancelled. Until the
operation is terminal, the endpoint does not support a final success or failure
label for the replay's write-success outcome. To count run B as a separate
unit, the protocol must use a separate namespace or begin B only after A's
operation is terminal and the shared state has been reset. Otherwise, the
analysis must represent the connection or group A and B as one unit. The
acknowledgement records acceptance of the request. Application of
the write is observed separately.

Function wrappers, process monitors, transaction logs, and environment
instrumentation can populate the record. Agent reports may identify effects
first visible in the trajectory, but an empty report does not show that no
relevant route exists.

\end{document}